\documentclass{article}
\usepackage{spconf,amsmath,epsfig}
\usepackage{multirow}
\usepackage{algorithm}
\usepackage{subfigure}
\usepackage{algorithmic}
\usepackage{amsthm, amsmath}
\usepackage{mathrsfs}
\usepackage{amssymb}

\let\OLDthebibliography\thebibliography
\renewcommand\thebibliography[1]{
  \OLDthebibliography{#1}
  \setlength{\parskip}{0pt}
  \setlength{\itemsep}{0pt plus 0.3ex}
}

\begin{document}\sloppy

\def\x{{\mathbf x}}
\def\L{{\cal L}}

\title{Annotation Efficient Person Re-Identification with Diverse Cluster-Based Pair Selection}
\name{Lantian Xue, Yixiong Zou, Peixi Peng, Yonghong Tian, Tiejun Huang}
\address{}

\maketitle

\begin{abstract}

Person Re-identification (Re-ID) has attracted great attention due to its promising real-world applications. 
However, in practice, it is always costly to annotate the training data to train a Re-ID model, 
and it still remains challenging to reduce the annotation cost while maintaining the performance for the Re-ID task. 
To solve this problem, we propose the Annotation Efficient Person Re-Identification method to select image pairs from an alternative pair set according to the fallibility and diversity of pairs, and train the Re-ID model based on the annotation.
Specifically, we design an annotation and training framework to firstly reduce the size of the alternative pair set by clustering all images considering the locality of features, secondly select images pairs from intra-/inter-cluster samples for human to annotate, thirdly re-assign clusters according to the annotation, and finally train the model with the re-assigned clusters.
During the pair selection, we seek for valuable pairs according to pairs' fallibility and diversity, which includes an intra-cluster criterion to construct image pairs with the most chaotic samples and the representative samples within clusters, an inter-cluster criterion to construct image pairs between clusters based on the second-order Wasserstein distance, and a diversity criterion for cluster-based pair selection.
Combining all criteria above, a greedy strategy is develop to solve the pair selection problem and select pairs.
Finally, the above clustering-selecting-annotating-reassigning-training procedure will be repeated until the annotation budget is reached.
Extensive experiments on three widely adopted Re-ID datasets show that we can greatly reduce the annotation cost while achieving better performance compared with state-of-the-art works.
\end{abstract}
\section{Introduction}

Person Re-Identification (Re-ID)~\cite{liu2020semantics,xin2019semi,xin2019deep} has attracted great attention in the field of computer vision, which retrieves images of a person from non-overlapping cameras. 
However, the training of a Re-ID model always relies on a large number of annotated samples with high annotation cost~\cite{liu2019deep}.

To solve this problem, some works have been proposed. For example, works~\cite{liu2020semantics,xin2019semi,xin2019deep} based on semi-supervised learning only require part of the dataset to be annotated, but the annotation cost is still high and the performance is still limited. Works~\cite{hu2021cluster,liu2019deep} inspired by active learning select few image pairs for human to annotate, but the annotation cost may still be related to the number of images. Therefore, it still remains challenging to reduce the annotating cost while maintaining the performance for Re-ID.

To achieve this goal, in this paper we utilize the verification based annotation, which verifies whether two images of each pair are from the same identity~\cite{hu2021cluster,liu2019deep}. Intuitively, valuable image pairs should be selected for annotation. Inspired by \cite{wu2020multi}, the value of a image pair can be defined as its informativeness, which could be divided into fallibility and diversity. 

Therefore, our aim is to select pairs from \textbf{the alternative pair set} according to the \textbf{fallibility} and \textbf{diversity} with an affordable budget, and train a Re-ID model based on the annotation. Observing this aim, we can find there are three important terms to design: 

(1) \textbf{The alternative pair set}, which means the set of all possible image pairs, influences the upper bound of the number of pairs to annotate. In view that images closer in the embedding space locally will tend to have the same semantic identity, we design to reduce the size of the alternative pair set by first clustering all images, then representing each cluster with a single representative image, and finally constructing pairs between representative images and other images. With the clustering operation, the size of the alternative pair set is related to the number of clusters instead of images.

(2) \textbf{Fallibility}, which means samples misclassified by the model (feature extractor), could be more informative for each selected pair.
Considering each cluster of images can be coarsely understood as an identity that the model classifies, the error that the model makes mainly lies in two aspects:

$ \bullet $ Within each cluster, images not belonging to the same identity are grouped into the same cluster, which we term as the chaotic samples. To select pairs from this situation, we propose to construct image pairs between the representative sample and the most chaotic sample of each cluster, which we term as the \textbf{intra-cluster criterion}.

$ \bullet $ Between clusters, clusters referring to the same identity are separated. To select pairs from this situation, we propose to construct image pairs between representative samples of all clusters based on the second order Wasserstein distance between clusters, termed as the \textbf{inter-cluster criterion}.

(3) \textbf{Diversity}. As samples from the same cluster tend to have similar appearances, it measures the information provided by all selected pairs. Intuitively, if all selected pairs are related to a few portion of clusters, the redundancy of the annotated pairs would be high. 
We select pairs according to KL-Divergence, aiming to average the frequency of each cluster for annotation, which we term as the \textbf{diversity criterion}.

Combining all criteria above, a greedy strategy is developed to find image pairs from the reduced alternative pair set.
After the annotation, for the intra-cluster situation, if images of a pair are verified to be from two identities, the cluster will be re-assigned to be separated into two clusters. In the meanwhile, for the inter-cluster situation, if images of a pair are annotated to belong to the same identity, the corresponding clusters will be re-assigned to be merged into one cluster.
Finally, each (re-assigned) cluster will be assigned with a pseudo-class label, which will be utilized to train the model. The above clustering-selecting-annotating-reassigning-training procedure will be repeated until the annotation budget is reached.

To sum up, our contribution can be summarized as:
    
$\bullet$ An annotation and training framework is proposed to annotate pairs from intra-/inter-cluster samples and train the model with re-assigned clusters, which greatly reduces the annotation cost while achieving better performance compared with current state-of-the-art annotation efficient methods.

$\bullet$ A pair selection strategy is proposed to select the most valuable pairs according to pair fallibility and diversity, and a greedy strategy is developed to solve the selection problem.

$\bullet$ Extensive experiments on three widely adopted Re-ID datasets validated the effectiveness of the proposed method.

\vspace{-0.4cm}
\section{Related Work}
\vspace{-0.2cm}

\textbf{Person Re-Id with less labeled data}:

~~Recently, some works try to use less labeled data to train the Re-ID model.
For example,
\cite{wang2018humanintheloop} learns a distance metric according to annotators' feed-backs. 
DRAL~\cite{liu2019deep} proposed to balance the data annotating with model training by a human-in-the-loop active learning manners. 
ARR~\cite{xu2021rethinking} proposed to select samples with high uncertainty. 
Compare with them, our diversity criterion is applied in cluster level with pairwise annotations, instead of the sample level measure.
To better deal with the redundancy, MASS~\cite{hu2021cluster} introduced clustering procedure with assigned pseudo labels and a scatter step to mine and annotate wrongly clustered samples. However, the pair selections and annotations remains within the cluster so there may still be some redundancy.

\noindent\textbf{Unsupervised Re-ID}:
Some works~\cite{zeng2020hierarchical,wang2020unsupervised,wu2019unsupervised} apply its own individual as identity for each sample. Apart from them, the cluster pseudo-labeling strategy~\cite{zhai2020ad,ge2020mutual,ge2020self} is proven to be effective. C-contrast~\cite{dai2021cluster} also clusters and assigns pseudo labels in the unsupervised setting but it directly updates cluster memory by single query when training. 
\cite{zeng2020energy} proposed an unsupervised clustering method focusing on generating clusters based on the energy distance and cluster deviations. Some works such as \cite{zeng2020hierarchical} also merge clusters in Re-ID.
Although with similar terms, our method differs in that we are searching for the most valuable pairs for annotating, which jointly optimizes for the fallibility and diversity.

\begin{figure*}[t]
	\centerline{\includegraphics[width=1.8\columnwidth,height=0.56\columnwidth]{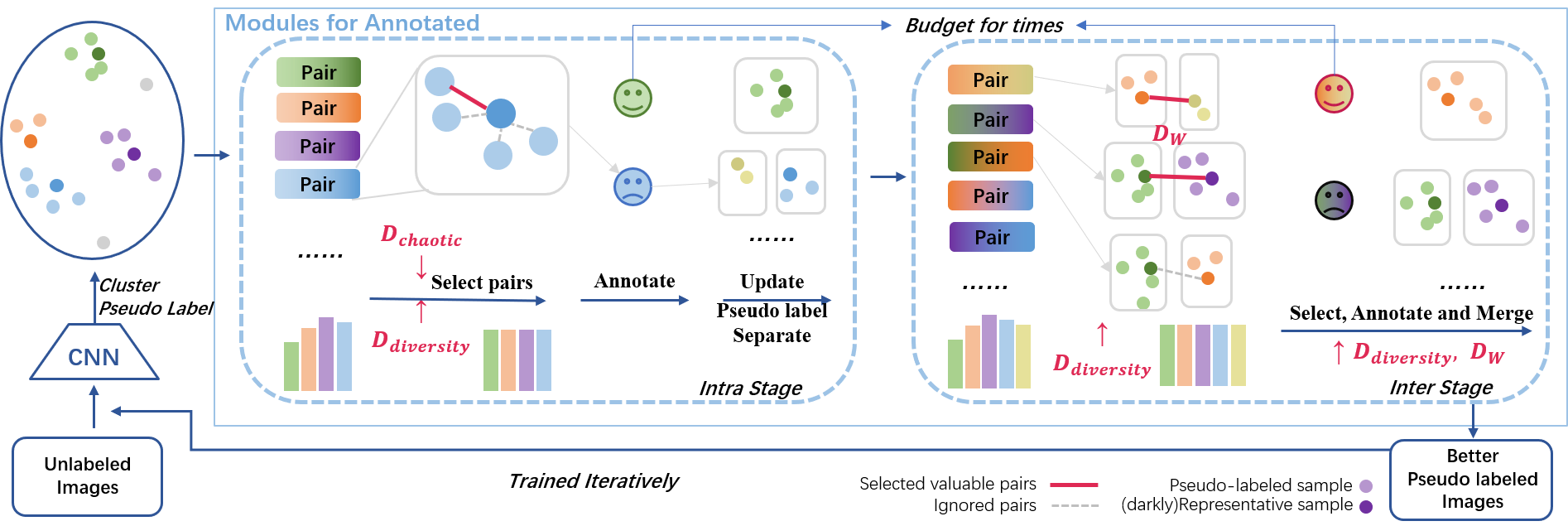}}
	\vspace{-0.2cm}
	\caption{
		Framework. We first cluster unlabeled images to reduce the size of the alternative pair set, then image pairs are selected according to the intra-cluster criterion, the inter-cluster criterion and the diversity criterion. After the annotation of selected pairs, clusters are re-assigned accordingly. Finally, the model will be trained with pseudo-class labels assigned for each cluster. The clustering-selecting-annotating-reassigning-training procedure will be repeated until the annotation budget is reached.}
	\label{fig:illustration}
	\vspace{-0.3cm}
\end{figure*}

\vspace{-0.4cm}
\section{Method}
\vspace{-0.3cm}

The propose the annotation efficient method includes procedures of clustering, pair-selection, annotation, cluster re-assignment, and training, as shown in Fig.~\ref{fig:illustration}.

\vspace{-0.3cm}
\subsection{Preliminaries}
\vspace{-0.1cm}

Given a person Re-ID dataset $D$, ordinarily it is divided into a training set $D_{train}$ and a test set $D_{test}$ with non-overlapping classes~\cite{Zheng_2015_ICCV,Zheng_2017_ICCV}, 
where each class refers to a person identity, and totally there are $M$ identities of $N$ person images in $D_{train}$.
For testing, $D_{test}$ is composed of queries and galleries for image-retrieval-like evaluation~\cite{Zheng_2015_ICCV,Zheng_2017_ICCV}. 

In this work, we follow \cite{liu2019deep,hu2021cluster} to start with the unsupervised Re-ID setting to assume we only have access to the unlabeled training samples, denoted as $D^{un}_{train}$, then aim to select data for human to annotate.
The annotation unit, as is adopted in current works~\cite{liu2019deep}, is the verification of a pair of images (e.g., $x_i$ and $x_k$) 
, which is marked as the verification function $v(x_i,x_k) \to \{0,1\}$, where 1 denotes they are from the same identity.
A deep network $F(x;\theta)$ with input $x$ and parameters $\theta$ will be used for embedding extraction. 

The pairs selected for annotation compose the selected pair set $\mathbb{P}_{sel} = \{..., (x_i, x_j), ...\}$ with a pre-defined maximum budget $T$ (verification times).
By repeating the selecting-annotating-training procedure for few iterations, the final trained model will be obtained, which will be evaluated on the test set.
Following \cite{liu2019deep,hu2021cluster}, the goal of our proposed method is to maximize the performance of the model $F(x;\theta)$ while reducing the annotation cost as much as possible. 

\vspace{-0.4cm}
\subsection{Optimization Objective}
\vspace{-0.2cm}

Intuitively, the most valuable pairs need to be selected for annotation. According to active learning~\cite{wu2020multi,liu2020pair}, the value of an image pair is defined as its informativeness, which can be divided into fallibility and diversity. The former item helps the model train more effectively, while the latter makes the model more robust. Thus we formulate the pair-selection task as an optimization problem, written as
\vspace{-0.1cm}
\begin{equation}
\begin{aligned}
\mathop{\max} O(\mathbb{P}_{sel}) &= Fallibility + \alpha Diversity \\
s.t. \lvert \mathbb{P}_{sel} \rvert &= T,  \mathbb{P}_{sel} \in \mathbb{P}_{total} = {D^{un}_{train}}^2
\label{con:optim_total}
\end{aligned}
\end{equation}
where $\alpha$ is a hyper-parameter, and $T$ is the budget. This means we select a set of pairs $\mathbb{P}_{sel}$ that provide the most fallibility and diversity overall from an alternative pair set $\mathbb{P}_{total}$.

Next, we will discuss how to simplify, specify, and solve it in our framework for the person Re-ID task.

\vspace{-0.4cm}
\subsection{Annotation and Training Framework}
\vspace{-0.2cm}
 
The alternative pair set $\mathbb{P}_{total}$ means the set of all possible image pairs. In the worst case, the size of $\mathbb{P}_{total}$ could be as large as N $\times$ M if we naively annotating all possible pairs. However, great redundancy exists in this naive strategy. 
Considering that samples closer in the embedding space locally will tend to have the same semantic identity, we may not need to verify every pair.
Therefore, same as ~\cite{hu2021cluster}, we first reduce the size of the alternative pair set by clustering all images.

In view that the clustering and pseudo-labeling based unsupervised learning methods~\cite{dai2021cluster} shows good performance in the Re-ID task, we consider to utilize this training procedure to take advantage of its simplicity and effectiveness. 

Specifically, given an unlabeled training set, we first cluster all samples through each feature $f_i = F(x_i;\theta)$ to form the cluster set $\mathcal{C}=\{c^1,c^2,...,c^{\lvert C \rvert}\}$, where the $j^{th}$ cluster is denoted as $c^j$. 
Later, the pair set $\mathbb{P}_{sel}$ will be selected based on these clusters, and clusters will be re-assigned according to the annotation on $\mathbb{P}_{sel}$, which will be illustrated in section~\ref{sec: pair selection} and \ref{sec: diversity}.
After the pair-selection and cluster-reassignment, all samples $x_i^j \in c^j$ will be re-assigned with a pseudo-label as $\widetilde{y_i} = j$, which will then be utilized to train $F(x;\theta)$. 
As the non-parametric classification loss has gained attention in person Re-ID recently, it inspires us to use 
the ClusterNCE~\cite{dai2021cluster} structure with its cluster based momentum memory update strategy. 
The loss function is written as
\vspace{-0.2cm}
\begin{equation}
L_i = -log\frac{exp(( f_i \cdot f^{\widetilde{y_i}} )/\tau)}{\sum_{j=0}^{\lvert C \rvert}exp((f_i \cdot f^j)/\tau)}
\label{con:cmloss}
\vspace{-0.2cm}
\end{equation}
where $f^j$ for cluster $j$ is initialized by averaging $f_i (x_i \in c^j)$ and updated by momentum during each iteration~\cite{dai2021cluster}.

\vspace{-0.3cm}
\subsection{Informative Pair Selection}
\vspace{-0.1cm}
\label{sec: pair selection}
After reducing the size of the alternative pair set $\mathbb{P}_{total}$, here we focus on the optimization objective in Eq.~\ref{con:optim_total}. We first introduce the fallibility term of the objective in this section, and then introduce the diversity term in section~\ref{sec: diversity}.

The CNN model $F(\cdot;\theta)$ is able to extract discriminative features after training. On that basis, we expect the annotation results of $\mathbb{P}_{sel}$ can further make up for deficiencies and correct mistakes. Intuitively, the more mistakes that $\mathbb{P}_{sel}$ can correct, approximately the more information it can provide for model training, and the more valuable they are. Therefore, two kinds of mistakes are set for pair selection in different stages:

$\bullet$ \textbf{intra-cluster stage}: correct mistakes that samples with different identities are clustered together. 

$\bullet$ \textbf{inter-cluster stage}: correct mistakes that samples from the same identity are clustered into two groups. 
Considering the annotation budget $T$, with the clustering operation, we first pick up a sample $r^j$ to represent the cluster $c^j$. Specifically, we select the sample closest to the cluster center $f_{mean}^j$ to be the representative sample, which is
\vspace{-0.1cm}
\begin{equation}
r^j = \mathop{\arg\min}_{x_i^j} \lVert f_i^j - f_{mean}^j\rVert_2, x_i^j\in c^j
\vspace{-0.2cm}
\end{equation}
For each stage, we construct a subset $\mathbb{P}_{stage} \in \mathbb{P}_{total}$ to be the alternative pair set for the model to select pairs, written as 
\begin{equation}
\mathbb{P}_{stage}=\left\{
\begin{array}{rcl}
\{(r^a, r^b)~|~c^a, c^b \in C\},&   {Intra-cluster}\\
\{(r^a, s^a)~|~c^a \in C \},~~~~&   {Inter-cluster}\\
\end{array} \right.
\end{equation}
where $s^a$ of $c^a$ will be described later.

Then, in the proposed annotation and training framework, for a pair in $\mathbb{P}_{stage}$, we use the possibility that it may cause cluster re-assignment to approximate its information. 
\vspace{-0.3cm}
\subsubsection{Intra-cluster stage: the chaotic degree distance}
\vspace{-0.1cm}
In the intra-cluster stage, the fallibility comes from whether a given sample belongs to the given cluster.
For each sample $x_i \in c^j$, a typical way to measure whether it is from another cluster (i.e., the chaotic degree) is to measure the distance between $f_{i}$ and $f_{mean}^j$. As the vertical position information is important in Re-ID, we also take this into consideration. In particular, before the final pooling layer of the CNN, we record each channel about which vertical portion the maximum value comes from, denoted as $g_i$. Thus, the chaotic degree of $x_i$ is defined as
\vspace{-0.1cm}
\begin{equation}
d_{intra}(x_i^j) =  \lVert f_i^j - f_{mean}^j\rVert_2 + \gamma \lVert g_i^j - g_{mean}^j\rVert_2, x_i^j\in c^j \label{con:chaotic}
\end{equation}
where $\gamma$ is a parameter for adjusting the proportions. 
This would be the intra-cluster criterion for the pair selection, which forms the first of fallibility in Eq.~\ref{con:optim_total} for the intra-cluster stage.
The chaotic distance of sample $s^j$ is then be recorded as the chaotic degree of this pair.

\vspace{-0.3cm}
\subsubsection{Inter-cluster stage: the Wasserstein distance}

In the inter-cluster stage, the fallibility comes from whether two clusters actually represent the same identity.
Intuitively, the closer two clusters are, the more likely that they might be from the same identity. Therefore, we would need the measuring of the distance between clusters to express the information provided by pair $(r^a, r^b)$.
Assuming each cluster as a Gaussian distribution, we use the Wasserstein distance~\cite{olkin1982distance} to measure the distance of every two clusters $c^a$ and $c^b$ as
\begin{small}
\begin{align}
-d^2_{inter}(c^a, c^b)  = 
-\mathcal{W}_{2} 
(\mathcal{N}(f_{mean}^a,\Sigma ^a);\mathcal{N}(f_{mean}^b, \Sigma ^b))^2 
\nonumber
=\\
\lVert f_{mean}^a- f_{mean}^b\rVert_2^2 + Tr({\Sigma^a+\Sigma^b-2({\Sigma^a}^{1/2} \Sigma^b {\Sigma^a}^{1/2})^{1/2}}) \label{con:w-distance}
\vspace{-0.5cm}
\end{align}
\end{small}
where the mean $f_{mean}^a$ and the variance $\Sigma^a$ come from embedding $f_i^a$ of all images $x_i \in c^a$, and every dimensions of $f$ are considered to be independent from each other. 

\vspace{-0.3cm}
\subsubsection{Cluster re-assignment}
In the intra-cluster stage, once annotators observed  $v(r^a,x^a_i) = 0$, i.e. $ y_{r^a} \neq y_{x^a_i}$, we separate them into two clusters and divide rest samples in $c^a$ according to their distance. In the inter-cluster stage, if $v(r^a,r^b) = 1$ i.e. $ y_{r^a} = y_{r^b}$, we set $ \widetilde{y}^a = \widetilde{y}^b = \widetilde{y}^a_{r^j}$. 
\vspace{-0.3cm}
\subsection{Diversity Criterion}
\vspace{-0.1cm}
\label{sec: diversity}

Considering one extreme situation that all selected pairs are relevant to one representative sample $r^a$, as the information provided by $r^a$ is limited, this would result in the redundancy in $\mathbb{P}_{sel}$.
Therefore, we also strive for the diversity of selected pairs.
Intuitively, we expect that when annotating, annotators have an equal probability of concerning each cluster. Therefore, we measure the diversity as the KL-divergence between the frequency of each cluster in selected pair set $\mathbb{P}_{sel}$ and the uniform distribution, termed as the diversity criterion for pair selection.
Let $p_j$ denotes the probability of $c^j$ in $\mathbb{P}_{sel}$ as
\vspace{-0.2cm}
\begin{equation}
p_j = \frac{cnt(c^j)}{T \times 2}
\label{con:pj}
\end{equation}
where $cnt(c^j)$ counts the times that $x^j \in c^j$ shows in $\mathbb{P}_{sel}$. 
So the KL-divergence between $p_j$ and the uniform distribution $U(0,1)$ is measured as
\vspace{-0.5cm}
\begin{equation}
D_{KL}(Q\lVert P) = \sum_{j=1}^{\lvert \mathop{C} \rvert} q \times (\log{\frac{q}{p_j}}) \label{con:kldis}
\vspace{-0.3cm}
\end{equation}
where Q is a uniform distribution.

\begin{table*}
	\small
	\centering
	\begin{tabular}{ccccccccccc}
		\hline
		\multicolumn{2}{c}{\multirow{2}{*}{Method}} & \multicolumn{4}{c}{Market1501}     & \multicolumn{4}{c}{DukeMTMC-reID} \\
		
		~&~& mAP   & Rank-1    & Rank-5    & Rank-10     & mAP   & Rank-1  & Rank-5    & Rank-10 \\
		\hline
		\multirow{4}{*}{Un}&HCT~\cite{zeng2020hierarchical}
		& 56.4    & 80.0    & 91.6    & 95.2    
		& 50.7    & 69.6    & 83.4    & 87.4   \\
		~&MMCL~\cite{wang2020unsupervised}
		& 45.5    & 80.3    & 89.4    & 92.3    
		& 40.2    & 65.2    & 75.9    & 80.0   \\
		~&SPCL(Un)~\cite{ge2020self}
		& 73.1    & 88.1    & 95.1    & 97.0   
		& -    & -    & -    & -   \\
		~ &C-contrast/ResNet50~\cite{dai2021cluster}
		& 82.6    & 93.0    & 97.0    & 98.1   
		& 72.8    & 85.7    & 92.0    & 93.5   \\
		\hline
		\multirow{3}{*}{UDA} &AD-C~\cite{zhai2020ad}
		& 68.3    & 86.7    & 94.4    & 96.5    
		& 54.1    & 72.6    & 82.5    & 85.5   \\
		
		~ &MMT~\cite{ge2020mutual}
		& 71.2    & 87.7    & 94.9    & 96.9   
		& 65.1    & 78.0    & 88.8    & \textbf{92.5}   \\

		~ &JVTC+~\cite{li2020joint}
		& 67.2    & 86.8    & 95.2    & 97.1   
		& 66.5    & 80.4    & 89.9    & 92.2   \\

		\hline

		\multirow{2}{*}{Semi}&SGCDPL~\cite{liu2020semantics}
		& 76.4    & 91.1    & -    & -   
		& 66.5    & 82.2    & -    & -   \\
		
		~ &MTML~\cite{zhu2019intra}
		& 65.2    & 85.3    & 96.2    & 97.6    
		& 50.7    & 71.7    & 86.9    & 89.6   \\

		\hline
		\multirow{2}{*}{Active} &DRAL~\cite{liu2019deep}
		& 66.3    & 84.2    & 94.3    & 96.6   
		& 56.0    & 74.3    & 84.8    & 88.4   \\
		
		~ &MASS~\cite{hu2021cluster}
		& 81.7    & 93.5    & 97.9    & 98.6  
		& 72.9    & 86.1    & 93.9    & 95.9   \\
		
		\hline\hline

		\multirow{3}{*}{Active}&Ours-baseline-GEMpooling    & 81.6   & 91.6   & 96.7   & 97.9
		& 70.1   & 83.2   & 91.0   & 93.2   \\
		~&Ours-Maxpooling    & {82.8}   & {92.5}   & 97.2   & 98.0
		& 74.5   & 86.4   & 92.8   & 94.7   \\
		~&Ours-GEMpooling    & \textbf{85.6}   & \textbf{93.6}   & \textbf{97.7}   & \textbf{98.4}   
		& \textbf{75.3}   & \textbf{86.9}   & \textbf{92.8}   & \textbf{94.7}   \\
		\hline
	\end{tabular}
	\vspace{-0.2cm}
	\caption{Compare with state-of-the-art method. The comparison of the annotation cost is included in section~\ref{sec: annotation cost}.}
	\label{tab:stoa}
	\vspace{-0.4cm}
\end{table*}

\vspace{-0.3cm}
\subsection{Solving Stage Optimization Objective}
\vspace{-0.1cm}
So far, we've simplified and specified Ep.~\ref{con:optim_total}. The final optimization objective function is written as
\vspace{-0.1cm}
\begin{equation}
\begin{aligned}
\min O(\mathbb{P}_{sel}) &= D + \alpha D_{KL} = \Sigma d + \alpha D_{KL} \\
s.t. \lvert \mathbb{P}_{sel} \rvert &= T,  \mathbb{P}_{sel} \in \mathbb{P}_{stage} 
\label{con:optimfunction}
\vspace{-0.4cm}
\end{aligned}
\end{equation}
where $D$ denotes the specific distance measuring for each stage, i.e., for the intra-cluster stage, $D$ means the chaotic distance, while for the inter-cluster stage, $D$ means the Wasserstein distance.
To solve this objective, however, it is hard to get the accurate optimal solution. 

Therefore, we use a greedy strategy to obtain a near-optimum solution, which selects a pair and pass to the human annotator once a time, according to which pair minimizes the increment of $O$ the most at each step, until the upper bound budget of annotation budget $T$ is reached. For each pair ($x^a, x^b$), the increment of $O$ if this pair is selected is represented as
\vspace{-0.2cm}
\begin{align}
o_{inc}(x^a,x^b) = d(x^a,x^b) + \frac{\alpha}{T}  o_{inc:D(KL)(x^a,x^b)}
\vspace{-0.2cm}
\label{con:oint}
\end{align}
Because $q$ is a constant as we fix $T$, 

$o_{inc:D(KL)}$ is only related to the change of $\log{cnt(c^j)}$ at each step. For example, if pair($x^a,x^b$) is selected at step $t$, then we have
\begin{small}
\begin{align}
\vspace{-0.2cm}
\begin{split}
o_{inc:D(KL)}(x^a,x^b) = &\beta  [\log{cnt_{t-1}(c^a)} - \log{cnt_{t}(c^a)}\\&
+\log{cnt_{t-1}(c^b)} - \log{cnt_{t}(c^b)}] 
\label{con:kldis_transform2}
\vspace{-0.2cm}
\end{split}
\end{align}
\end{small}
where $\beta$ is also a constant calculated by $q$, $\alpha$ and $T$.
The $cnt(c^a)$ will increase by 1 or 2 once the pair that includes samples from $c^a$ is selected at each step $t$.
Because of the diversity criterion we use, selections are relatively sparse, so one pair has limited effect on others, and the greedy strategy is sufficient to get a near-optimum solution.

\vspace{-0.4cm}
\subsection{Iterative Model Training}
\vspace{-0.1cm}
By iterative executing the clustering-selecting-annotation-reassigning-training procedure for iterations, $F(x;\theta)$ will be finally obtained. We distributed the annotation budget to each epoch in a Gaussian manner, with the mean and variance set to epoches/2. So there are more annotation times assigned in the middle of training. This is because at the beginning, the discriminability of the model $F(x;\theta)$ is limited and valuable pairs are hard to select. On the other hand, at the end, $F(x;\theta)$ has become stable and annotations cannot be fully utilized. 

\vspace{-0.2cm}
\section{Experiments}
\vspace{-0.2cm}

\subsection{Datasets and Evaluation Protocols}
\vspace{-0.2cm}

The Market-1501~\cite{Zheng_2015_ICCV} dataset contains 32668 images of 1501 identities.
DukeMTMC-ReID~\cite{Zheng_2017_ICCV}, which is a subset of DukeMTMC~\cite{ristani2016performance}, contains 36411 images of 1404 identities in total.
The MSMT17\cite{wei2018person} dataset contains 4101 identities with 126,441 images.
We use the Cumulative Matching Characteristic (CMC) curve and the mean average precision (mAP) to evaluate the proposed method.
\vspace{-0.5cm}
\subsection{Implementation Details}
\vspace{-0.2cm}
ResNet-50~\cite{he2016deep} which produces 2048-d features is adopted as the CNN model followed by a GEM pooling~\cite{gu2018attention}, a BatchNorm~\cite{ioffe2015batch} layer and L2-Normalization.
Batch size is set to 256 while each identity contains 16 instances.
The input images will be transformed with flipping and random erasing~\cite{zhong2020random}. 
We use Adam~\cite{kingma2014adam} as the optimizer with weight decay 0.0005. The learning rate is set to 0.00035. 
Following the baseline method ~\cite{dai2021cluster}, DBSCAN~\cite{ester1996density} is chosen with $\epsilon$ set to 0.4/0.7 and the minimum number of points required to form a dense region is set to 4 with image Jaccard distance~\cite{zhong2017re}.

\vspace{-0.3cm}
\subsection{Results and Comparisons with State-of-the-Arts}
\vspace{-0.2cm}

\begin{table}
	\small
	\centering
	\begin{tabular}{ccccc}
		\hline
		Method  & mAP & Rank-1 & Rank-5  &  Rank-10 \\
		\hline
		(active) MASS~\cite{hu2021cluster}   & 30.0   & 54.1   & 65.4   & 70.4   \\
		(Un) C-Contrast~\cite{dai2021cluster}     & 33.3   & 63.3   & 73.7   & 77.8 \\
		Ours-baseline      & 30.7   & 59.0   & 70.7   & 75.3   \\
		Ours            & \textbf{35.5}   & \textbf{63.4}   & \textbf{74.5}   & \textbf{79.0}   \\

		\hline
	\end{tabular}
	\vspace{-0.2cm}
	\caption{Compare with state-of-the-art method on MSMT17.}
	\vspace{-0.2cm}
	\label{tab:sotamsmt}
\end{table}

\begin{table}
	\small
	\centering
	\begin{tabular}{ccccc}
		\hline
		Method  & mAP & Rank-1 & Rank-5  &  Rank-10 \\
		\hline
		Ours $w/o$ intra.  & 83.2   & 93.4   & 97.4   & 98.5   \\
		Ours $w/o$ inter.  & 82.5   & 92.1   & 96.9   & 98.0   \\
		Ours $w/o$ diver.  & 84.9   & 93.5   & 97.6   & 98.5   \\
		\hline
		Ours  & 85.6   & 93.6   & 97.7   & 98.4   \\
		\hline
		Ours-upper      & 87.5   & 94.6   & 97.7   & 98.7   \\
		
		\hline
	\end{tabular}
	\vspace{-0.2cm}
	\caption{Ablation studies: methods with different modules and the fully-supervised upper bounds on Market1501. Our method is close to the upper bound.}
	\label{tab:ablation-module}
	\vspace{-0.2cm}
\end{table}

\begin{figure}[t]
	\centerline{\includegraphics[width=0.98\columnwidth]{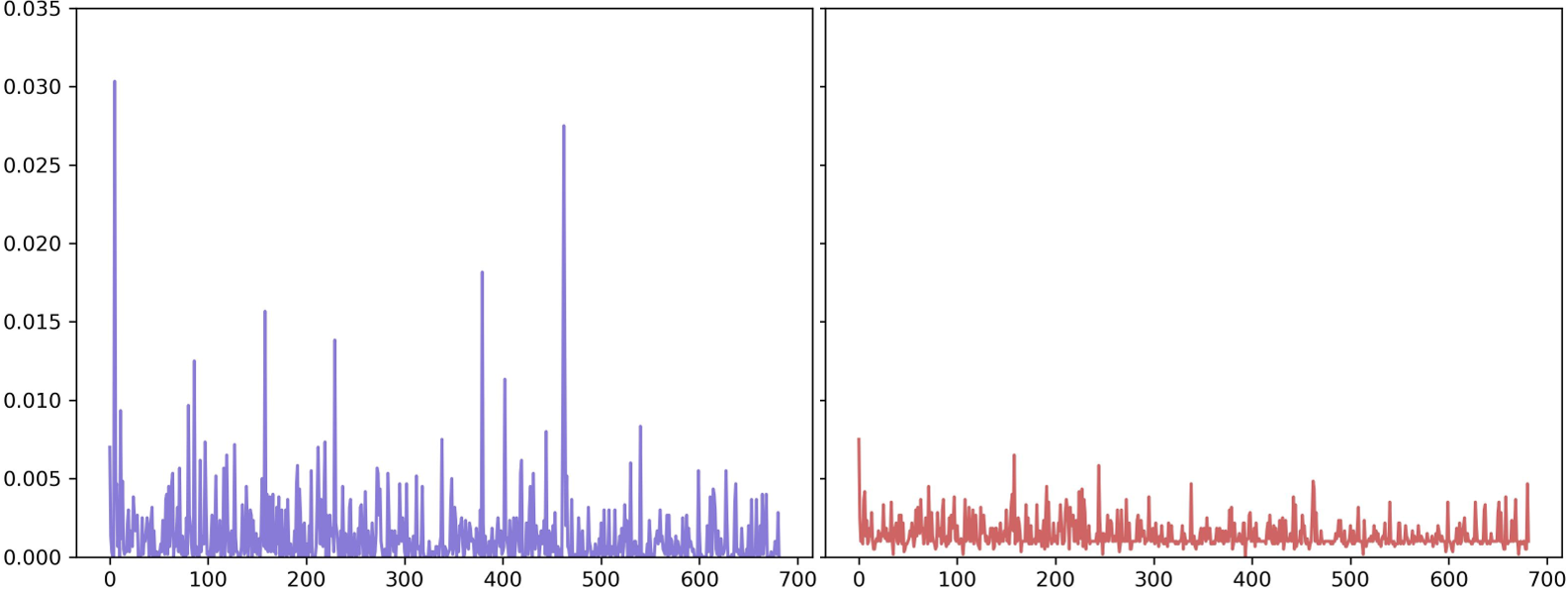}}
	\vspace{-0.2cm}
	\caption{Visualization of $p_j$ with (right) and without (left) the diversity criterion. }
	\label{fig:diver}
	\vspace{-0.1cm}
\end{figure}

Results show in Table~\ref{tab:stoa} and \ref{tab:sotamsmt}, with $T$ set to 3000. The baseline method does not need annotations.
Unsupervised methods are marked as 'Un', unsupervised person re-id methods are marked as 'UDA'. Our method can outperform them all with simpler structure design.
Compared with semi-supervised methods. Our proposal also shows good performance even with less annotations.
Compare with the active learning based method DRAL~\cite{liu2019deep,hu2021cluster}, our method also outperforms them, with much less annotation cost.

\vspace{-0.4cm}
\subsection{Ablation Studies}

\vspace{-0.2cm}
\subsubsection{\textbf{Ablation of modules}} 
\vspace{-0.2cm}
\label{sec:abmodule}
The ablation study is in Table \ref{tab:ablation-module}.
`Ours $w/o$ \textit{intra}' and `Ours $w/o$ \textit{inter}' show separately the Inter/Intra pair selecting and pseudo label updating strategy both improve the performance.
Moreover, by replacing pseudo labels with ground-truth labels, we report the upper bound performance of our method, which trains the model with full supervision. We can see that \textbf{the performance of our model is quite close to `Ours-upper'}, which verifies our aim: reducing the annotation cost while maintaining the performance.
We also discuss the effect of the diversity regularized selection through Figure~\ref{fig:diver}. It shows the values of $p_j$ defined from Eq.~\ref{con:pj} of each representative sample $r^j$. Through our KL-divergence regularized criterion, $p_j$ will be more uniform, which means the annotator looks at each sample with a relatively uniform probability.

\begin{table}
	\small
	\centering
	\vspace{-0.2cm}
	\begin{tabular}{ccccc}
		\hline
		method & $T$  & mAP & Rank-1 & Rank-5  \\
		\hline
		DRAL~\cite{liu2019deep} & $10N \approx 120k$ & 66.3 & 84.2 & 94.3  \\
		MASS~\cite{hu2021cluster} & $5N \approx 60k$ & 81.7  & 93.5  & 97.9  \\
		\hline
		
		 Ours  &1000    & 84.1  & 93.3  & 97.4   \\
		&3000    & \textbf{85.6}  & \textbf{93.6}  & \textbf{97.7}   \\
		
		\hline
		
	\end{tabular}
	\vspace{-0.2cm}
	\caption{mAP and Rank-1 with different $T$ on Market1501. We can greatly reduce the annotation cost while achieving better performance.}
	\vspace{-0.5cm}
	\label{tab:hyperparameter}
\end{table}

\vspace{-0.4cm}
\subsection{Effects of Annotates and Annotate Costs}
\vspace{-0.2cm}
\label{sec: annotation cost}

Table~\ref{tab:hyperparameter} shows the performance with different budget $T$. We can see that with a larger $T$, the model performs better, which is consistent with the intuition that more supervision leads to better results. 
The total annotation cost of our proposal in all epochs $T$ is set to $3000$. 
We compare our annotation cost with MASS~\cite{hu2021cluster} and DRAL~\cite{liu2019deep}, which is the state-of-the-art methods for reducing the annotation cost for the Re-ID task.
$N \times 5$ in \cite{hu2021cluster} is reported to be the annotation cost, and $N \times 10$ in \cite{hu2021cluster} is the adopted budget. Our budget is $T < N < N \times 5 < N \times 10$, meaning that we use a much lower annotation cost to achieve better results.

\vspace{-0.2cm}
\section{Conclusion}
\vspace{-0.2cm}
To reduce the annotation cost while maintaining the performance for the Re-ID task, in this paper we 
propose the Annotation Efficient Person Re-Identification method to select image pairs from an alternative pair set according to the fallibility and diversity of pairs, and train the Re-ID model based on the annotation. Extensive experiments show that we can achieve better performance with much less annotations compared with current works.
\vspace{-0.3cm}

\small
\bibliographystyle{IEEEbib}
\bibliography{icme2022template}

\end{document}